\documentclass[10pt,twocolumn,a4paper]{esaAI}
\usepackage[acronym]{glossaries}
\usepackage[detect-all]{siunitx}

\newcommand{\reffig}[1]{Fig.~\ref{#1}}
\newcommand{\reftab}[1]{Table~\ref{#1}}
\newcommand{\refsec}[1]{Sec.~\ref{#1}}
\newcommand{\fone}{$F_1$-score}

\title{Camera-Pose Robust Crater Detection from Chang'e 5}

\def\authorEmail{matthew.rodda@adelaide.edu.au}

\author[1]{Matthew Rodda\thanks{Corresponding author. E-Mail: \authorEmail}}
\author[1]{Sofia McLeod}
\author[1]{Ky Cuong Pham}
\author[1]{Tat-Jun Chin}
\affil[1]{The University of Adelaide, Adelaide, Australia}

\begin{document}

\newacronym{cda}{CDA}{crater-detection algorithm}
\newacronym{ce5}{CE5}{Chang'e 5}
\newacronym{dem}{DEM}{digital-elevation map}
\newacronym{dtm}{DTM}{digital-terrain map}
\newacronym{p}{P}{precision}
\newacronym{r}{R}{recall}
\newacronym{mrcnn}{$M_{RCNN}$}{Mask R-CNN}
\newacronym{ercnn}{$E_{RCNN}$}{Ellipse R-CNN}
\newacronym{lro}{LRO}{Lunar Reconnaissance Orbiter}
\newacronym{nac}{NAC}{Narrow-Angle Camera}

\makeCustomtitle

\begin{abstract}
As space missions aim to explore increasingly hazardous terrain, accurate and timely position estimates are required to ensure safe navigation.
Vision-based navigation achieves this goal through correlating impact craters visible through onboard imagery with a known database to estimate a craft’s pose.
However, existing literature has not sufficiently evaluated \gls{cda} performance from imagery containing off-nadir view angles.
In this work, we evaluate the performance of Mask R-CNN for crater detection, comparing models pretrained on simulated data containing off-nadir view angles and to pretraining on real-lunar images. 
We demonstrate pretraining on real-lunar images is superior despite the lack of images containing off-nadir view angles, achieving detection performance of 63.1\% \fone{} and ellipse-regression performance of 0.701 intersection over union.
This work provides the first quantitative analysis of performance of \glspl{cda} on images containing off-nadir view angles.
Towards the development of increasingly robust \glspl{cda}, we additionally provide the first annotated \gls{cda} dataset with off-nadir view angles from the Chang'e 5 Landing Camera, available here: \url{https://zenodo.org/doi/10.5281/zenodo.11326449}.

\end{abstract}

\glsresetall

\section{Introduction}
\begin{figure}[h]
    \centering
    \includegraphics[width=.99\columnwidth]{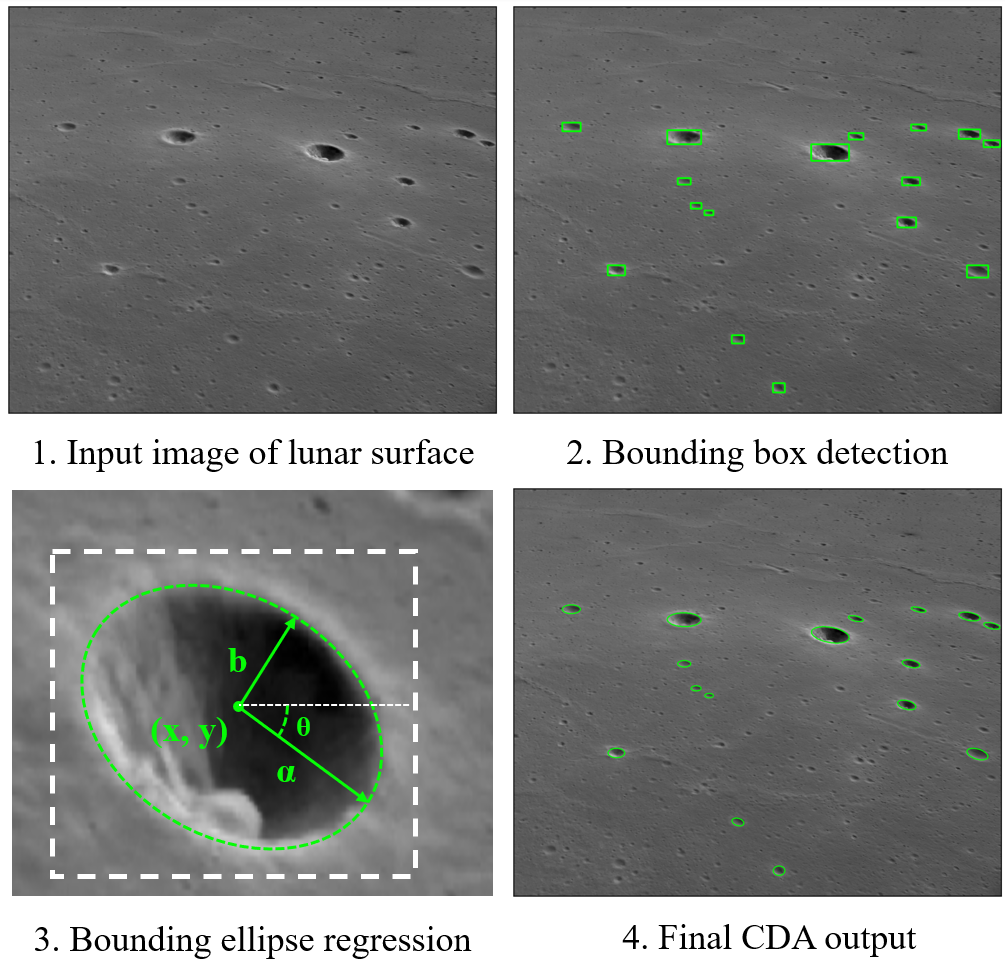}
    \caption{Steps of \gls{cda} pipeline, demonstrating bounding box detection and ellipse regression on image from the Chang'e 5 landing camera~\cite{CE5}.}
    \label{fig:model_architectures}
\end{figure}

Accurate pose estimation of spacecraft is increasingly valuable as future missions aim to hazardous yet scientifically valuable terrain, such as the lunar south pole.
Recent investigations into vision-based navigation systems have utilized camera measurements cross-referenced with onboard maps to provide accurate estimations of pose \cite{Christian2021-ts, Maass2020}.
A crucial initial step for vision-based navigation is detection of salient features from camera measurements \cite{Christian2021-ts}.
For planetary bodies without thick atmospheres (eg. the Moon and asteroids), impact craters provide a valuable source of detectable and persistent landmarks visible on the surface \cite{Maass2016}.
Techniques to detect such impact craters from imagery are named \glspl{cda}, with a pipeline shown in \reffig{fig:model_architectures}.

Existing literature explores \gls{cda} performance in a narrow operating scenario, often using a \gls{dem} or strictly nadir-pointing images as input.
In this work, we provide the first analysis of \gls{cda} performance on real optical imagery with off-nadir view angles on a newly annotated dataset from the Chang'e 5 landing camera.

\section{Related Works}
There exists significant literature describing the creation and evaluation of \glspl{cda} under different operating conditions~\cite{tewari2023, Woicke2018-lg, DeLatte2019-ye, Salamuniccar2008-oy}.
Typically, these algorithms aim to increase the completeness of crater catalogues on planetary bodies such as the Moon or Mars by automating detection \cite{Benedix2020, Lee2021, AliDib2020}.
As such, many \glspl{cda} use \glspl{dem} as input, as this data is free from variation due to lighting or view angle \cite{tewari2023}.
However, for optical navigation, crater detection from a \gls{dem} is constraining as these maps require many measurements to be accurately defined~\cite{LRO-Dataset}.
Hence, for optical navigation, it is preferable to detect craters from optical imagery.

\glspl{cda} using real optical imagery are relatively less explored.
Yang et al.\ applied 7 \glspl{cda} to an optical imagery dataset collected from the \gls{lro} \gls{nac}, demonstrating state-of-the-art detection performance of 86.97\%$F_1$-score with a Faster R-CNN architecture~\cite{CraterDaNet}.
Mao et al.\ applied a Dual U-Net architecture to optical imagery collected from the \gls{lro} Wide-Angle Camera, in addition to a \gls{dem}, demonstrating that this additional data source leads to with improved performance as compared to 4 \glspl{cda} using only the \gls{dem}~\cite{DualUNet}.
While these \glspl{cda} predict from optical imagery, again their relevance to optical navigation is questionable as images are reconstructed mosaics and strictly nadir-pointing~\cite{Robinson2010}.

The only analysis of \gls{cda} performance with off-nadir view angles is provided by Maas et al.\, using a novel segmentation-based \gls{cda} on both simulated images of the near-earth asteroid Eros in 2016 and real images from Chang'e 3 orbiter in 2020 \cite{Maass2016, Maass2020}.
While both datasets contain significant off-nadir view angles, a quantitative analysis of \gls{cda} performance was not performed.

\subsection{Contributions}

In this work, we describe the annotation of the first \gls{cda} dataset with significant off-nadir view angles, using images collected from the Chang'e 5 landing camera \cite{CE5}.
Addressing off-nadir view angles constitutes a significant advancement to existing analyses as these view angles can obscure already complex surface morphology \cite{Giannakis2024}.
We analyze the performance of \gls{mrcnn}, a previously high-performing \gls{cda} from literature, after pretraining on real and simulated \gls{cda} datasets from literature~\cite{AliDib2020, he2018mask}.
We demonstrate that, despite the lack of off-nadir view angles, datasets containing real lunar images achieve best detection and ellipse-regression performance after finetuning.

\section{Datasets}
\label{sec:datasets}
\subsection{CE5-CDA}
CE5-CDA is a newly annotated dataset of real lunar images collected by the Chang'e 5 Landing Camera~\cite{CE5}.
132 images were hand-labelled by expert annotators, describing craters present within the image with a bounding ellipse of the crater rim.
The first 100 images of the Chang'e 5 landing camera were labelled to create a training set, and every tenth image of the remaining 313 were labelled to create the testing set.

CE5-CDA, like many \gls{cda} datasets, provides an incomplete annotation of visible craters.
This is due to the number of small craters visible in early images, with the first 150 images containing >300 visible craters.
Annotators prioritised annotating the largest craters for labelling first, then craters with a well-defined rim.
On average, the dataset contains 50 labelled craters per image, with later images within the dataset containing fewer labelled craters, as fewer craters are visible as the lander approaches the lunar surface.

To our knowledge, CE5-CDA is both the first annotated \gls{cda} dataset of real-lunar images containing off-nadir view angles, and the first annotated dataset of the Chang'e 5 landing camera.
Hence, as part of this work, we make the CE5-CDA dataset available to the research community for further analysis.

\subsection{CRESENT}
CRESENT is a dataset of rendered images of the lunar surface, with crater rims supervised by a bounding ellipse~\cite{McLeod2024}.
Images were rendered from 100km above the surface, with off-nadir view angles ranging between 0 to 60 degrees, in increments of 10 degrees, mimicking the expected orbital conditions of a lunar surveillance mission~\cite{McLeod2024}.

This simulated dataset was valuable as camera pose was known exactly.
This allowed automated supervision of the rendered images by projecting Robbin's crater catalogue onto the surface and analysis of \gls{cda} performance as off-nadir view angles varied~\cite{LDEM, Robbins2019}.

Notably, as no complete lunar-crater catalogue exists, annotation of craters present within each image is incomplete.
As such, many CRESENT images contain multiple unlabelled instances of craters that appear obvious to the human-eye.
The subset of CRESENT images within 0-75E, 0-30N was used for practical purposes, with the number of images per off-nadir view angle shown in \reftab{tab:simulated_images}.
A 20\% random stratified split of images from each off-nadir view angle was selected as a validation set.

\begin{table}[h]\renewcommand{\arraystretch}{1.2}
\begin{center}
\begin{tabular}{c || c | c | c | c | c | c | c } 
\hline\hline
Angle & 0$^{\circ}$ & 10$^{\circ}$ & 20$^{\circ}$ & 30$^{\circ}$ & 40$^{\circ}$ & 50$^{\circ}$ & 60$^{\circ}$ \\ \hline \hline
Images & 461 & 450 & 437 & 415 & 381 & 325 & 225 \\
\hline \hline
\end{tabular}
\caption{Count of images per off-nadir view angle.}
\label{tab:simulated_images}
\end{center}
\end{table}

\subsection{LRO-NAC}
LRO-NAC is a manually supervised dataset of 20 images from the \gls{lro} \gls{nac}, containing over 20 000 annotated craters \cite{CraterDaNet}.
While the LRO-NAC is the smallest dataset utilized in this work, it is a valuable source of real lunar image mosaics, containing images under a range of lighting conditions.
Similarly to CE5-CDA and CRESENT, LRO-NAC is incomplete, however it provides the most complete annotation of visible craters of the three datasets, with hundreds of annotations per image.
The dataset consists of 12 800x800 resolution images used for training, and 8 1000x1000 resolution images used for validation.

\section{Methodology}
In this section, we describe the training and evaluation process of \gls{mrcnn}.
\gls{mrcnn} is a specialization of Faster R-CNN, a two-stage object detection architecture that has exhibited high performance across a large range of computer vision tasks, including state-of-the-art performance in crater detection \cite{ren2016faster, CraterDaNet}.
\gls{mrcnn} uses a specialized head to predict a segmentation mask, a binary mask with the same resolution as the input image, where masked pixels represent the presence of an object in the original image.
After training, the near-elliptical segmentation mask produced by \gls{mrcnn} is fit with an ellipse using a least-square fitting algorithm\cite{Pilu, OpenCV}.

\subsection{Training}
All training was performed in Pytorch with a stochastic gradient descent optimizer until validation loss plateaued, at which point the best performing weights were restored \cite{Pytorch}.
Training was performed on a g4dn.16xlarge AWS instance, with GPU acceleration using a Tesla T4.
Training time for both architectures were similar, taking $<3$ hours to train on CRESENT, $<40$ minutes on LROC-NAC and $<15$ minutes on CE5-CDA.

\subsection{Evaluation}
\gls{cda} performance was evaluated for detection and ellipse-regression separately.
Detection performance is evaluated using \gls{p}, \gls{r} and $F_1$-score, where true positive ($T_P$) predictions are determined by the intersection-over-union (IoU) of predicted and ground-truth bounding boxes being greater than 0.5.
Ellipse-regression performance of each $T_P$ is then evaluated, again using IoU.
Ellipse IoU ($E_{\text{IoU}}$) of a $T_P$ was approximated by projecting the ellipses onto an image at the same resolution of the input sample.
The intersection and union were then calculated at pixel resolution on the projected ellipses.

\section{Results}
\label{sec:results}
Detection and ellipse-regression performance of \gls{mrcnn} was evaluated after training on different subsets of the datasets described in \refsec{sec:datasets}.
Best detection and ellipse-regression performance is achieved by pretraining on LRO-NAC, before finetuning on the CE5-CDA dataset, as shown in \reftab{tab:change5_results}.
Despite CRESENT's number of training images and inclusion of images with off-nadir view angles, bridging the domain gap between the simulated images and real CE5-CDA images during finetuning was challenging.
In fact, training on CE5-CDA alone outperformed the model pretrained on CRESENT in both detection and ellipse-regression.

While LRO-NAC lacked images with off-nadir view angles, it supplied real-lunar images and the highest completeness of annotated craters, which may have resulted in its best performance.

\gls{cda} performance without finetuning was also evaluating, yielding poor results.
This suggests that existing datasets are not sufficient for generalizing to the real images and off-nadir view angles present in CE5-CDA.

\begin{table}[h]\renewcommand{\arraystretch}{1.2}
\begin{center}
\begin{tabular}{c || c | c | c | c | c } 
\hline\hline
Pretraining & P(\%) & R(\%) & F$_1$(\%) & $E_{\text{IoU}}$ & E($T_P$) \\ \hline\hline
CRESENT             & $  5.3 $ & $  4.4 $ & $  6.9 $ & $ 0.557 $ & $  1.3 $ \\ 
LRO-NAC             & $ 12.3 $ & $ 27.3 $ & $ 16.6 $ & $ 0.549 $ & $  8.0 $ \\ 
\textbf{None}       & $ 63.1 $ & $ 56.3 $ & $ 60.0 $ & $ 0.687 $ & $ 18.7 $ \\ 
\textbf{CRESENT}    & $ 52.4 $ & $ 55.3 $ & $ 54.3 $ & $ 0.674 $ & $ 18.3 $ \\ 
\textbf{LRO-NAC}    & $ 57.0 $ & $ 68.4 $ & $ 63.1 $ & $ 0.701 $ & $ 22.2 $ \\ 
\hline\hline
\end{tabular}
\caption{\gls{cda} performance on CE5-CDA test set. The pretraining column denotes what dataset, if any, was used to initially train the \gls{cda}, and bolding denotes whether the \gls{cda} was then finetuned on the CE5-CDA training set. The expected number of $T_P$ predictions is included as many pose estimation algorithms require a minimum of 5 to function.}
\label{tab:change5_results}
\end{center}
\end{table}

\begin{figure*}[!t]
\centering
\includegraphics[width=.99\textwidth]{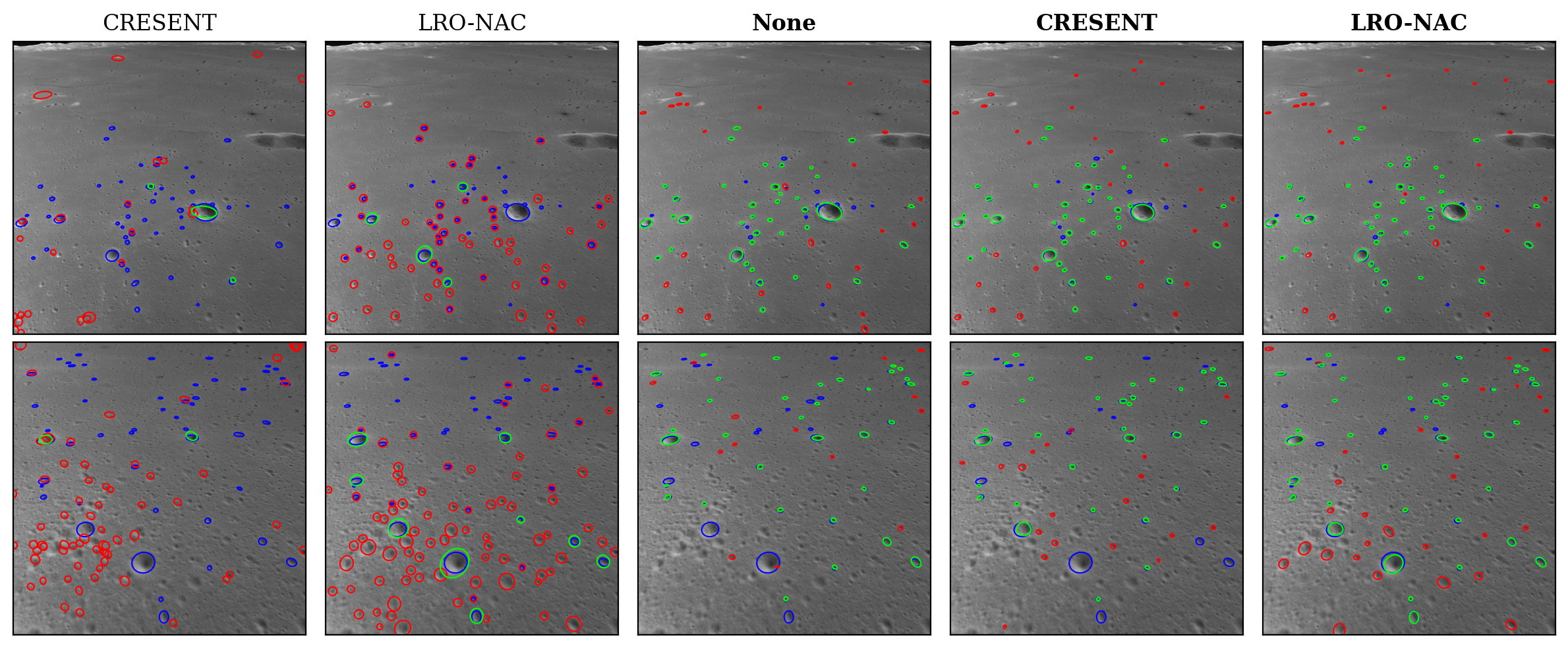}
\caption{
 Sample images from CE5-CDA, with \gls{cda}-regressed ellipses.
 Rows show two different samples from the CE5-CDA test set, while columns separate inference from each \gls{cda}.
 Blue ellipses denote ground truth, while green and red ellipses show true positive and false positive predictions respectively.
 }
\label{fig:case_study}
\end{figure*}

\reffig{fig:case_study} shows ellipse-regressions from each of the 5 \glspl{cda} evaluated.
Many false positive predictions visible in these samples appear to be real-unlabelled instances, demonstrating the affect of incompleteness in annotation.
While the incompleteness of CE5-CDA is a limitation of the dataset, it is representative of the incompleteness of real-lunar crater catalogues.
In operation, a \gls{cda} may produce detections of visible craters that are not catalogued, which may be considered false positives as they are useless of pose estimation.

\begin{figure}[h]
    \centering
    \includegraphics[width=.95\columnwidth]{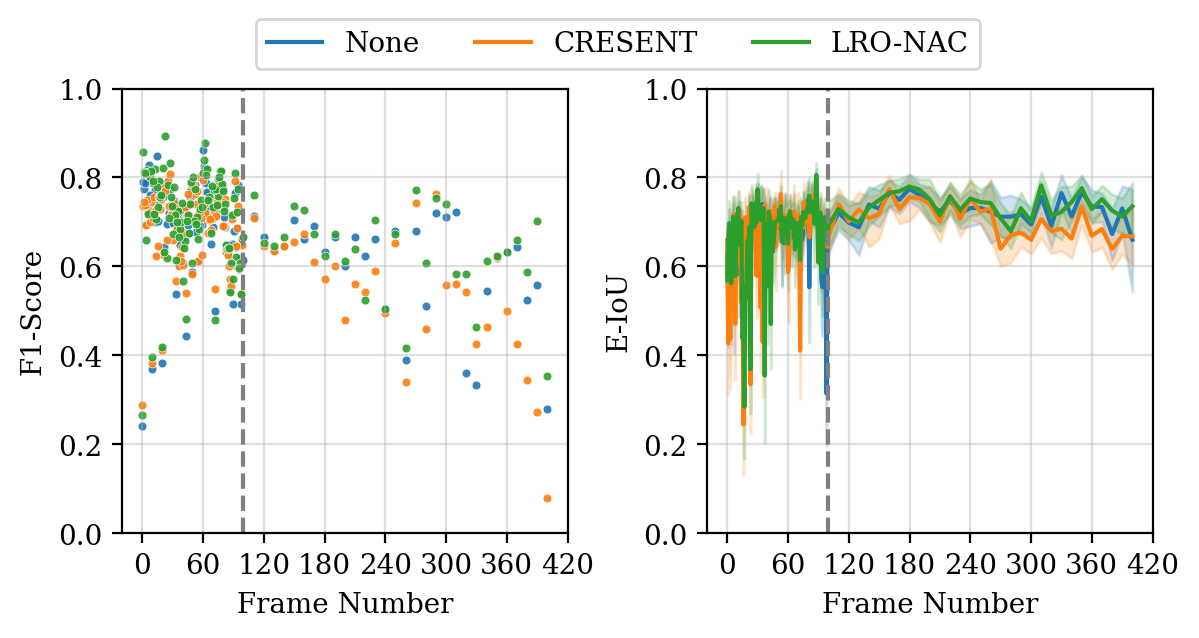}
	\caption{Reduction in detection and ellipse-regression performance over frame number of finetuned \glspl{cda}.
 The end of the training-set and beginning of the test-set is denoted with a grey line.}
	\label{fig:ce5_drift}
\end{figure}

As the Chang'e 5 lander approached the surface, the landing camera recorded images from both decreasing altitudes and off-nadir view angles \cite{CE5}.
\reffig{fig:ce5_drift} shows the \gls{cda} pretrained on LRO-NAC demonstrated the best generalization from the training set to this new scenario, in both detection and ellipse-regression.

 \section{Discussion}
In \refsec{sec:results}, we demonstrate that simulated images are not sufficient for pretraining a \gls{cda} despite a greater number of training samples and closer replication of operating conditions of the final task.
Future work may explore how domain adaptation could bridge the simulated-to-real domain gap to improve performance after finetuning on real lunar images.

Future work may additionally explore automated labelling of real lunar images by projecting known crater catalogues onto existing mosiacs, using annotated position information.
As demonstrated in this work, real lunar images are necessary for best \gls{cda} performance, however manual labelling of large datasets is infeasible.
Automating labelling of nadir-pointing image mosiacs could allow manual labelling efforts be focused towards annotating the previous Chang'e 3 and recently launched Chang'e 6 mission data, containing off-nadir view angles.

Future work may also investigate the affect of illumination conditions in tandem with camera-pose on \gls{cda} performance.
Such analysis could inform whether strategic collection of images in a power-constrained environment may result in best crater-detection performance and consequently higher quality pose estimations.

\section{Conclusion}
In this work, we provide the first quantitative analysis of \gls{cda} performance on real lunar images containing off-nadir view angles.
We demonstrate that an existing state-of-the-art \gls{cda} achieves poor performance in this operating scenario, but can be improved through finetuning.
While pretraining on simulated images containing off-nadir view angles yields no improvement, pretraining on real nadir-pointing lunar mosiacs improved both detection and ellipse-regression performance, achieving 63.1\% \fone{} $0.701 E_{\text{IoU}}$ respectively.
For development of robust \glspl{cda}, we show that inclusion of off-nadir training images is necessary for sufficient detection performance in this operating scenario.

\pagebreak

\printbibliography
\addcontentsline{toc}{section}{References}

\end{document}